\pdfoutput=1

\documentclass[letterpaper, 10pt, conference]{ieeeconf}      
 
\IEEEoverridecommandlockouts                              

\usepackage{multicol}
\usepackage{multirow}
\usepackage{algpseudocode}
\usepackage{color}
\usepackage{xcolor}
\usepackage{graphicx}
\usepackage{bm}
\usepackage{amsmath}
\usepackage{amssymb}
\usepackage{tabularx}
\usepackage{subcaption}
\usepackage{graphicx}
\usepackage{amsmath}
\usepackage{amssymb}
\usepackage{booktabs}
\usepackage{subcaption}
\usepackage{url}
\usepackage{cite}
\usepackage{prettyref}
\usepackage{booktabs}
\usepackage{siunitx}
\usepackage{comment}
\usepackage{booktabs}
\usepackage{hyperref}
\usepackage[capitalize]{cleveref}

\usepackage[table]{xcolor} 
\usepackage{colortbl}
\usepackage{tikz}
\usepackage[dvipsnames]{xcolor}

\usepackage{caption}
\usepackage{adjustbox} 
\newcommand{\emoji}[1]{\includegraphics[height=1em]{#1}}

\usepackage{lettrine}

\usepackage{tabularx, booktabs}
\usepackage{makecell} 
\usepackage{tabularx} 
\definecolor{Gray}{gray}{0.9}

\usepackage[all]{hypcap}
\usepackage[capitalize]{cleveref}
\crefname{section}{Sec.}{Secs.}
\Crefname{section}{Section}{Sections}
\Crefname{table}{Table}{Tables}
\crefname{table}{Tab.}{Tabs.}

\usepackage[ruled]{algorithm2e} 

\SetAlFnt{\small}
\SetAlCapFnt{\small}
\SetAlCapNameFnt{\small}
\SetAlCapHSkip{0pt}

\newrefformat{fig}{Figure~\ref{#1}}
\newrefformat{sec}{Section~\ref{#1}}
\newrefformat{tab}{Table~\ref{#1}}
\newrefformat{alg}{Algorithm~\ref{#1}}
\newrefformat{eqn}{Equation~\ref{#1}}

\usepackage{titlesec}
\titlespacing{\section}{0pt}{*1.4}{*0.7}
\titlespacing{\subsection}{0pt}{*1.1}{*0.4}

\hypersetup{
    colorlinks,
    linkcolor={blue!80!black},
    citecolor={blue!80!black},
    urlcolor={blue!100!black}
}

\makeatletter
\newcommand{\printfnsymbol}[1]{%
  \textsuperscript{\@fnsymbol{#1}}%
}
\makeatother
\newcommand{\sysname}{GaussianFormer3D}

\usepackage[dvipsnames]{xcolor}

\title{\huge \sysname:  Multi-Modal Gaussian-based Semantic Occupancy Prediction with 3D Deformable Attention}

\author{Lingjun Zhao, Sizhe Wei, James Hays and Lu Gan
\thanks{The authors are with the Georgia Institute of Technology, Atlanta, GA 30332, USA. L. Zhao is supported by IRIM Ph.D. Fellowship at Georgia Institute of Technology. Email: {\tt\footnotesize \{lzhao360,swei,hays,lgan\}@gatech.edu}.}
}

\begin{document}
\maketitle

\begin{abstract}
3D semantic occupancy prediction is essential for achieving safe, reliable autonomous driving and robotic navigation. Compared to camera-only perception systems, multi-modal pipelines, especially LiDAR-camera fusion methods, can produce more accurate and fine-grained predictions. Although voxel-based scene representations are widely used for semantic occupancy prediction, 3D Gaussians have emerged as a continuous and significantly more compact alternative. In this work, we propose a multi-modal Gaussian-based semantic occupancy prediction framework utilizing 3D deformable attention, namely GaussianFormer3D. We introduce a voxel-to-Gaussian initialization strategy that provides 3D Gaussians with accurate geometry priors from LiDAR data, and design a LiDAR-guided 3D deformable attention mechanism to refine these Gaussians using LiDAR-camera fusion features in a lifted 3D space. Extensive experiments on real-world on-road and off-road autonomous driving datasets demonstrate that GaussianFormer3D achieves state-of-the-art prediction performance with reduced memory consumption and improved efficiency. Project website: \href{https://lunarlab-gatech.github.io/GaussianFormer3D/}{https://lunarlab-gatech.github.io/GaussianFormer3D/}.
\end{abstract}


\section{Introduction}

Perception systems are essential for the development of safe, reliable and intelligent autonomous vehicles and field robots~\cite{zhang2024visionsurvey}. Among various perception tasks, 3D semantic occupancy prediction is particularly crucial as it enables fine-grained understanding of both geometric and semantic information of the environments~\cite{sscnet}. For autonomous driving (AD), recent advances in vision-based occupancy prediction have shown impressive results on large-scale datasets~\cite{nuscenes, semantickitti}.
However, the sensitivity of cameras to lighting variations and their limited depth accuracy still underscore the need to incorporate additional sensing modalities for robust AD.

LiDAR sensors have been widely applied to AD for perception tasks such as 3D object detection~\cite{voxelnet, pointpillars}. Compared to cameras, LiDAR provides more accurate depth information and finer geometric relationships of objects, making it particularly advantageous for
3D semantic occupancy prediction~\cite{lmscnet, s3cnet, scpnet, motionsc, ssa-sc, ssc-rs}. However, LiDAR-based pipelines often struggle to capture accurate semantics for small objects, where camera-based methods excel~\cite{monoocc}.
To balance geometric accuracy and semantic richness, multi-modal fusion algorithms have been proposed to leverage the strengths of complementary sensing modalities. Current approaches include mainly LiDAR-camera fusion~\cite{coocc, occgen, occmamba, occfusiondepthfree} and camera–radar fusion~\cite{licrocc}, with LiDAR-camera fusion showing superior performance and becoming the dominant choice.

Most LiDAR-camera occupancy networks employ a 3D voxel-based~\cite{coocc, occgen, occmamba, occfusiondepthfree, slcfnet} representation to model a scene as a dense grid-based structure. Despite achieving comparable performance, these methods inevitably suffer from redundant empty grids and high computational costs. Recently, inspired by the success of 3D Gaussian splatting~\cite{3dgs}, an object-centric Gaussian representation has been explored in vision-based 3D semantic occupancy prediction, achieving improved computational efficiency. 
GaussianFormer series~\cite{gaussianformer, gaussianformer2} represent a scene as a set of 3D Gaussians, each consisting of a mean, covariance and semantics. These Gaussians are refined using a 2D deformable attention~\cite{deformabledetr}, and then processed by an efficient Gaussian-to-voxel splatting module to predict semantic occupancy.
Despite high efficiency, current Gaussian-based methods~\cite{gaussianformer, gaussianformer2} rely solely on 2D image to update 3D Gaussians, limiting their ability to
model 3D space with accurate depth and fine-grained geometric structure. How to effectively leverage other sensor modalities, such as LiDAR, to refine and obtain a more accurate 3D Gaussian representation for efficient semantic occupancy prediction remains unexplored and challenging.


\begin{figure}[t] 
    \centering
    \includegraphics[width=\columnwidth]{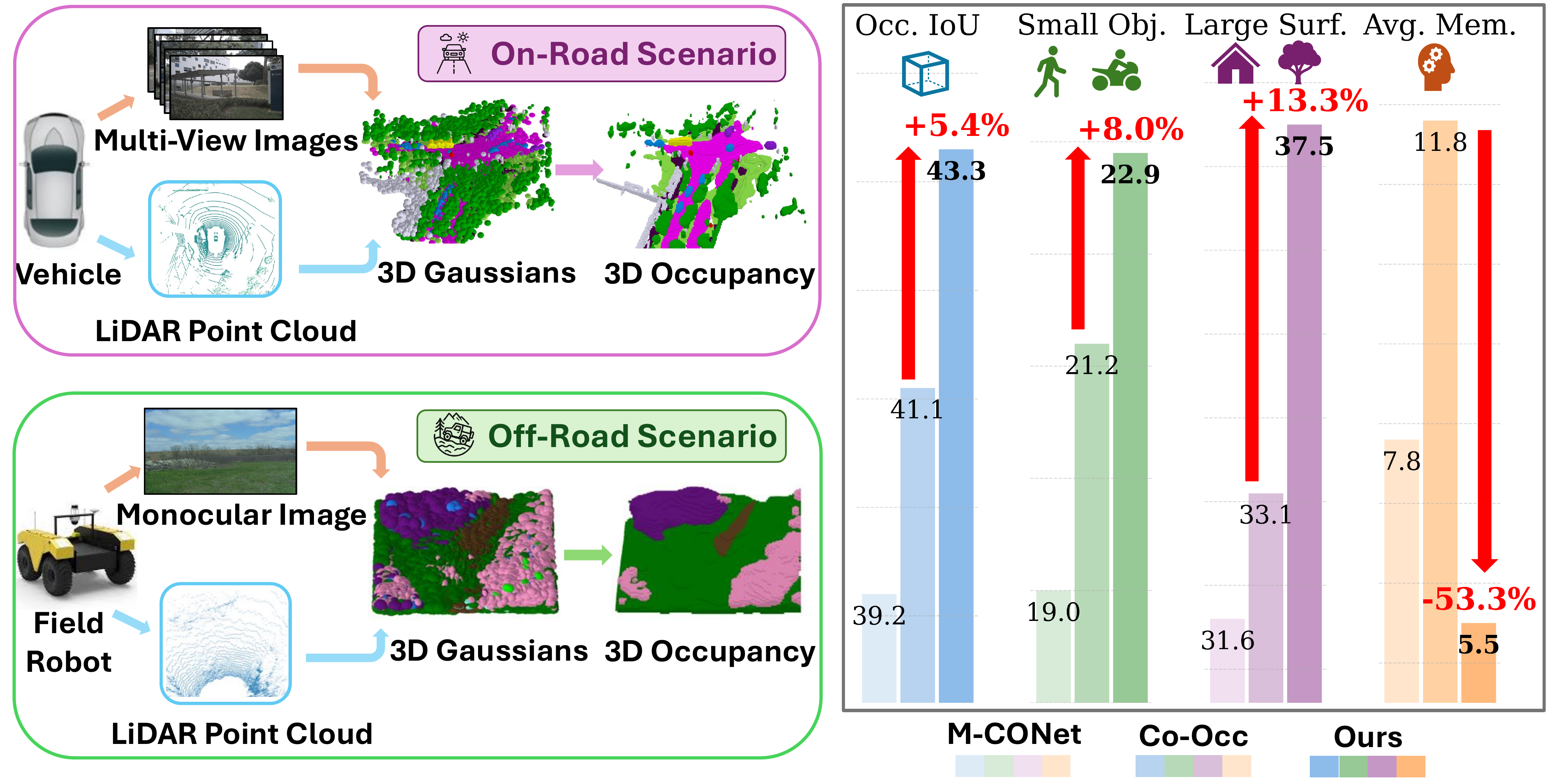}
    \vspace{-15pt}
    \caption{\textbf{We propose a new LiDAR-camera fusion-based semantic occupancy prediction framework using 3D Gaussians.} 
     We evaluate it on both on-road and off-road driving scenarios. Our method demonstrates superior performance on overall occupancy Intersection-of-Union (IoU), achieves substantial performance gains on small objects (\textit{pedestrian}, \textit{motorcycle}) and large surfaces (\textit{manmade}, \textit{vegetation}), and consumes less memory during inference.}
    \vspace{-15pt}
    \label{fig:pitch}
\end{figure}

In this work, we propose \textbf{GaussianFormer3D}: a multi-modal Gaussian-based semantic occupancy prediction framework with 3D deformable attention, as shown in \cref{fig:pitch}. GaussianFormer3D models a scene using 3D Gaussians initialized from LiDAR voxel features, updates Gaussians through 3D deformable attention in a LiDAR-camera unified 3D feature space, and predicts semantic occupancy via Gaussian-to-voxel splatting. To the best of our knowledge, our model is the first multi-modal semantic occupancy network that employs an object-centric Gaussian-based scene representation. In summary, our main contributions are as follows:
\begin{itemize}
    \item We propose a novel multi-modal Gaussian-based semantic occupancy prediction framework. By integrating LiDAR and camera data, ours significantly outperforms camera-only baselines with similar memory usage.
    \item We design a voxel-to-Gaussian initialization module to provide 3D Gaussians with geometry priors from LiDAR, and also develop an enhanced 3D deformable attention mechanism to update Gaussians by aggregating LiDAR-camera fusion features in a lifted 3D space. 
    \item We conduct extensive evaluations on two on-road datasets, nuScenes-SurroundOcc~\cite{surroundocc} and nuScenes-Occ3D~\cite{tian2024occ3d}, along with an off-road dataset, RELLIS3D-WildOcc~\cite{wildocc}. Results show that ours outperforms state-of-the-art dense grid-based methods while achieving reduced memory consumption and improved efficiency.
\end{itemize}

\section{Related Work}

\begin{figure*}[t] 
    \centering
    \includegraphics[height=7.5cm]{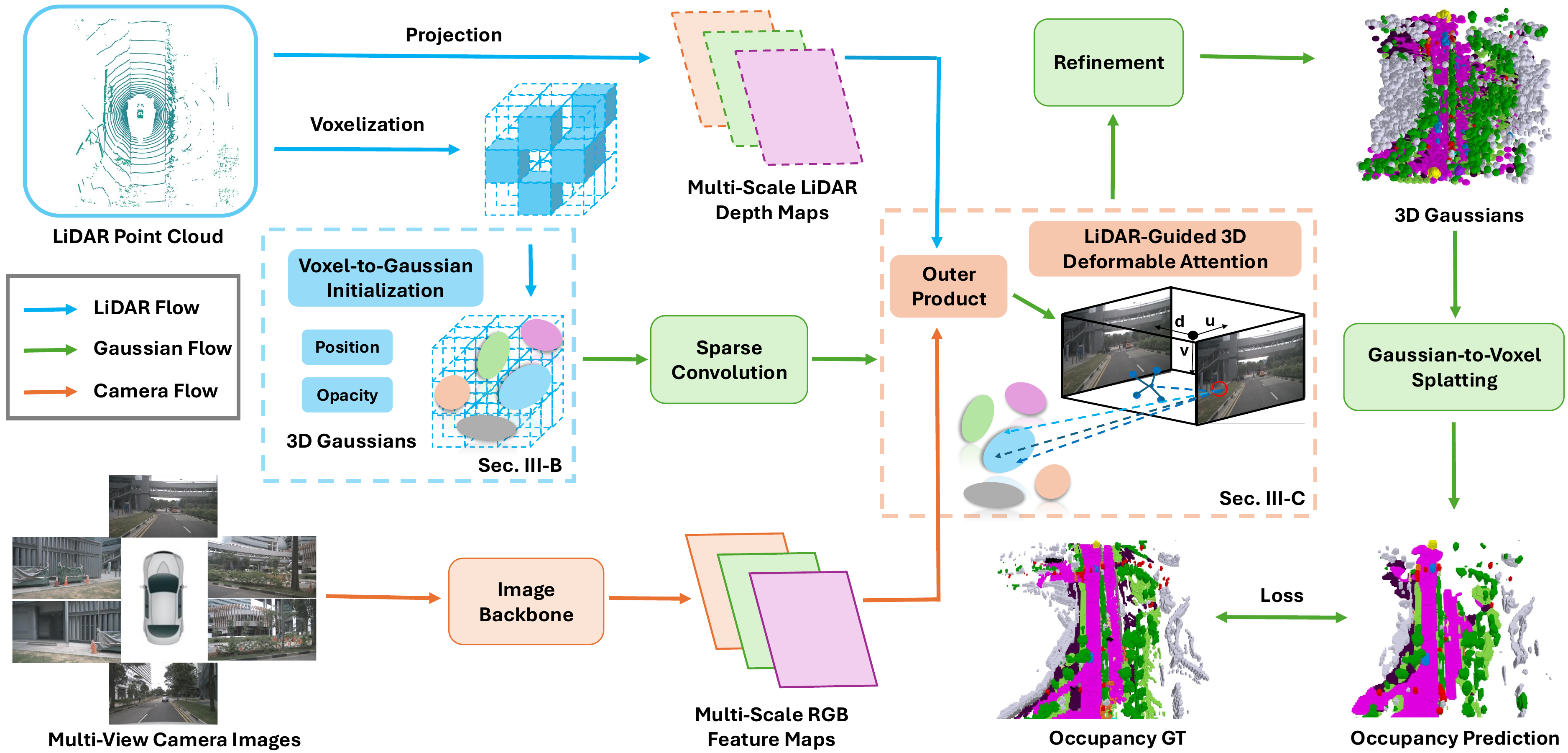}
    \vspace{-3pt}
    \caption{\textbf{GaussianFormer3D Overview.} 
    We first voxelize LiDAR point clouds to obtain non-empty voxel features for initializing the position and opacity of 3D Gaussians. Then LiDAR depth maps and camera feature maps are extracted respectively, and multiplied via outer product to construct a lifted 3D feature space. Gaussians are iteratively updated with sparse convolution, 3D deformable attention, and property refinement. Gaussians are eventually processed by a Gaussian-to-voxel splatting module to generate semantic occupancy.}
    \vspace{-15pt}
    \label{fig:overall}
\end{figure*}

\textbf{Multi-Modal Semantic Occupancy Prediction.} Multi-modal occupancy prediction methods generally outperform single-modal approaches, as different modalities provide complementary information. Among them, LiDAR-camera fusion is the top-performing configuration, combining LiDAR’s accurate depth and geometry sensing with the powerful semantic understanding capability of cameras. Similar to single-modal pipelines, most LiDAR-camera occupancy prediction networks are also voxel-based~\cite{coocc, occgen, occmamba, occfusiondepthfree, openoccupancy}.
CONet~\cite{openoccupancy} proposes a coarse-to-fine pipeline to sample 3D voxel features for refining the coarse occupancy prediction. 
Co-Occ~\cite{coocc} obtains multi-modal voxel features through a geometric and semantic-aware fusion module, and employs a NeRF-based implicit volume rendering regularization~\cite{nerf} to enhance the fused representation. 
OccGen~\cite{occgen} and OccMamba~\cite{occmamba} encode multi-modal inputs to produce voxel fusion features, and then decode the features using diffusion denoising and hierarchical Mamba modules, respectively. 
OccFusion~\cite{occfusiondepthfree} transforms LiDAR and camera inputs into multi-modal voxel features via 2D deformable attention~\cite{deformabledetr}.

\textbf{3D Gaussians for Autonomous Driving.}
Due to the inherent advantages of modeling scenes explicitly and continuously, 3D Gaussians~\cite{3dgs} have been adopted as the scene representation over the traditional grid-based solutions in 3D semantic occupancy prediction~\cite{gaussianformer, gaussianformer2, gaussianworld}.
3D Gaussians also demonstrated their superiority in real-time image rendering and novel view synthesis, and thus have been adopted for driving scene reconstruction and simulation~\cite{drivinggaussian, streetgaussians, splatad}. 
Furthermore, end-to-end autonomous driving~\cite{gaussianad} and visual pre-training~\cite{gaussianpretrain} utilize 3D Gaussians as the driving world representation for various downstream perception and planning tasks. 
However, these approaches are mainly designed for camera-only autonomous driving, neglecting the potential of multi-modal data in Gaussian initialization and updating. GSPR~\cite{gspr} proposes a Gaussian-based multi-modal place recognition algorithm, and SplatAD~\cite{splatad} designs the first 3D Gaussian splatting pipeline to render both LiDAR and camera data. In this work, we explore utilizing multi-modal data, especially from LiDAR and camera sensors, to learn a fine-grained 3D Gaussian representation for more accurate and efficient semantic occupancy prediction.

\section{Method}

The overview of GaussianFormer3D is presented in~\cref{fig:overall}.

\subsection{Scene as 3D Gaussian Representation} \label{sec:3.1}
Semantic occupancy prediction aims to jointly predict the semantic information and geometric structure of the scene. Given multi-view images \mbox{$\mathcal{I}=\{\mathbf{I}_i\}_{i=1}^{N_c}$} and LiDAR point cloud \mbox{$\mathcal{P}=\{\mathbf{P}_i\}_{i=1}^{N_p}$, $\mathbf{P}_i = (x_i, y_i, z_i, \eta_i)$} containing the 3D location and intensity of each point, the goal is to predict the semantic occupancy grid \mbox{$\mathbf{O}\in \mathcal{C}^{X\times Y\times Z}$}, where $N_c$, $N_p$, $\mathcal{C}$ denote the number of camera views, the number of LiDAR points, and the set of semantic classes, and \mbox{$X\times Y\times Z$} is the size of the voxel grid to be predicted. Unlike uniform grids in traditional grid-based representations, 3D Gaussians can adaptively represent the regions of interest due to the universal approximation capability of Gaussian mixtures~\cite{gaussianformer}. Specifically, a scene is modeled as a set of 3D Gaussians \mbox{$\mathcal{G}=\{\mathbf{G}_i\}_{i=1}^{N_g}$}, where $N_g$ is the total number of Gaussians in a scene. Each Gaussian $\mathbf{G}_i$ is parameterized by its mean \mbox{$\mathbf{m}_i \in \mathbb{R}^3$}, rotation \mbox{$\mathbf{r}_i \in \mathbb{R}^4$}, scale \mbox{$\mathbf{s}_i \in \mathbb{R}^3$}, opacity \mbox{$\sigma_i \in [0, 1]$} and semantic label \mbox{$\mathbf{c}_i \in \mathbb{R}^{|\mathcal{C}|}$}. The value of Gaussian $\mathbf{G}$ evaluated at location $\mathbf{x}$ can be calculated as:
\begin{align}
    \mathbf{g}(\mathbf{x}; \mathbf{G}) = \sigma\cdot{\rm{exp}}\big(-\frac{1}{2}(\mathbf{x}-\mathbf{m})^{\rm T} \mathbf{\Sigma}^{-1} (\mathbf{x}-\mathbf{m})\big)\mathbf{c},
    \label{eq: gaussian_distribution} \\
    \mathbf{\Sigma} = \mathbf{R}\mathbf{S}\mathbf{S}^{\rm T}\mathbf{R}^{\rm T}, \quad \mathbf{S} = {\rm{diag}}(\mathbf{s}), \quad \mathbf{R} = {\rm{q2r}}(\mathbf{r}),
\end{align}
where $\mathbf{\Sigma}$, $\mathbf{R}$ and $\mathbf{S}$ denote the covariance matrix, rotation matrix and scale matrix. $\rm{diag}(\cdot)$ is the diagonal matrix construction and $\rm{q2r}(\cdot)$ is the quaternion-to-rotation transformation. By summing the contributions of all Gaussians at location $\mathbf{x}$, the occupancy prediction can be formulated as:
\begin{align}
    \hat{\mathbf{o}}(\mathbf{x}; \mathcal{G})=\sum_{i=1}^{N_g}\mathbf{g}_i(\mathbf{x}; \mathbf{m}_i,\mathbf{s}_i,\mathbf{r}_i,\sigma_i,\mathbf{c}_i).
    \label{eq: weighted summation}
\end{align}
The Gaussian-to-voxel splatting module is designed to only aggregate Gaussians within the neighborhood of a targeted voxel instead of querying all Gaussians in a scene to improve efficiency and reduce unnecessary computation and storage~\cite{gaussianformer}. Thus, \cref{eq: weighted summation} can be furthur approximated by replacing $N_g$ with $N_g(\mathbf{x})$,
where $N_g(\mathbf{x})$ is the number of neighboring Gaussians at location $\mathbf{x}$. During training, the Gaussian-based occupancy model is trained in an end-to-end manner, supervised by the ground truth semantic occupancy label \mbox{$\mathbf{\bar{O}}\in \mathcal{C}^{X\times Y\times Z}$}. Both cross entropy loss $L_{ce}$ and the lovasz-softmax loss $L_{lov}$ are used for optimization.

\subsection{Voxel-to-Gaussian Initialization} \label{sec:3.2}
Two sets of 3D Gaussian features are adopted following GaussianFormer~\cite{gaussianformer}. The first set consists of learnable Gaussian physical properties \mbox{$\mathcal{G}=\{\mathbf{G}_i \in \mathbb{R}^d \}_{i=1}^{N_g}$} introduced in~\cref{sec:3.1}, where \mbox{$d=11+|\mathcal{C}|$}, which are also our learning targets.
The second set is non-learnable high-dimensional Gaussian features \mbox{$\mathcal{Q}=\{\mathbf{Q}_i \in \mathbb{R}^m \}_{i=1}^{N_g}$}, where $m$ is the feature dimension, serving as queries for the attention mechanism~\cite{transformer} and implicitly encoding the spatial and semantic information during the Gaussian update. Previous work~\cite{gaussianformer} randomly initializes the Gaussian physical properties, and optimizes these properties iteratively through multiple refinement modules. This design constrains Gaussians to learn complex 3D geometry information solely from 2D images, which inevitably encounters inaccurate spatial modeling.

To resolve this issue, we propose a LiDAR-based voxel-to-Gaussian initialization strategy to initialize the mean and opacity of Gaussians with geometry priors from LiDAR data, as indicated in the dashed blue box in \cref{fig:overall}. Specifically, we first aggregate the most recent $N_f$ LiDAR scans into a combined point cloud \mbox{$\mathcal{\bar P}=\{\mathcal{P}_i\}_{i=1}^{N_f}$}. Then we voxelize the combined point cloud and compute the feature of each non-empty voxel as the mean position and intensity of all points within it. These LiDAR-based voxel features are then used to initialize the position and opacity of 3D Gaussians:
\begin{equation}
    \mathbf{m}_i =  \frac{1}{|\mathcal{P}_v|}\sum_{j\in \mathcal{P}_v}(x_j, y_j, z_j), \quad \sigma_i = \frac{1}{|\mathcal{P}_v|}\sum_{j\in \mathcal{P}_v}\eta_j,
\end{equation}
where \mbox{$i \in \{1, ..., N_g\}$} denotes the index of Gaussians to be initialized and \mbox{$v \in \{1, ..., N_v\}$} denotes the index of all non-empty voxels; $\mathcal{P}_v$ is the set of LiDAR points in $\mathcal{\bar P}$ falling into voxel $v$. When \mbox{$N_g<N_v$}, we randomly choose a subset of $N_g$ non-empty voxels to initialize Gaussians, otherwise, a subset of $N_v$ Gaussians are randomly selected and initialized with non-empty voxels.  After initialization, we apply a 3D sparse convolution module to the initialized 3D Gaussians for self-encoding. The features and interactions of Gaussians are efficiently extracted and aggregated through the sparse convolution for updating the Gaussian queries.

\subsection{LiDAR-Guided 3D Deformable Attention} \label{sec:3.3}
\begin{figure}[!h]
    \centering
    \includegraphics[width=\linewidth]{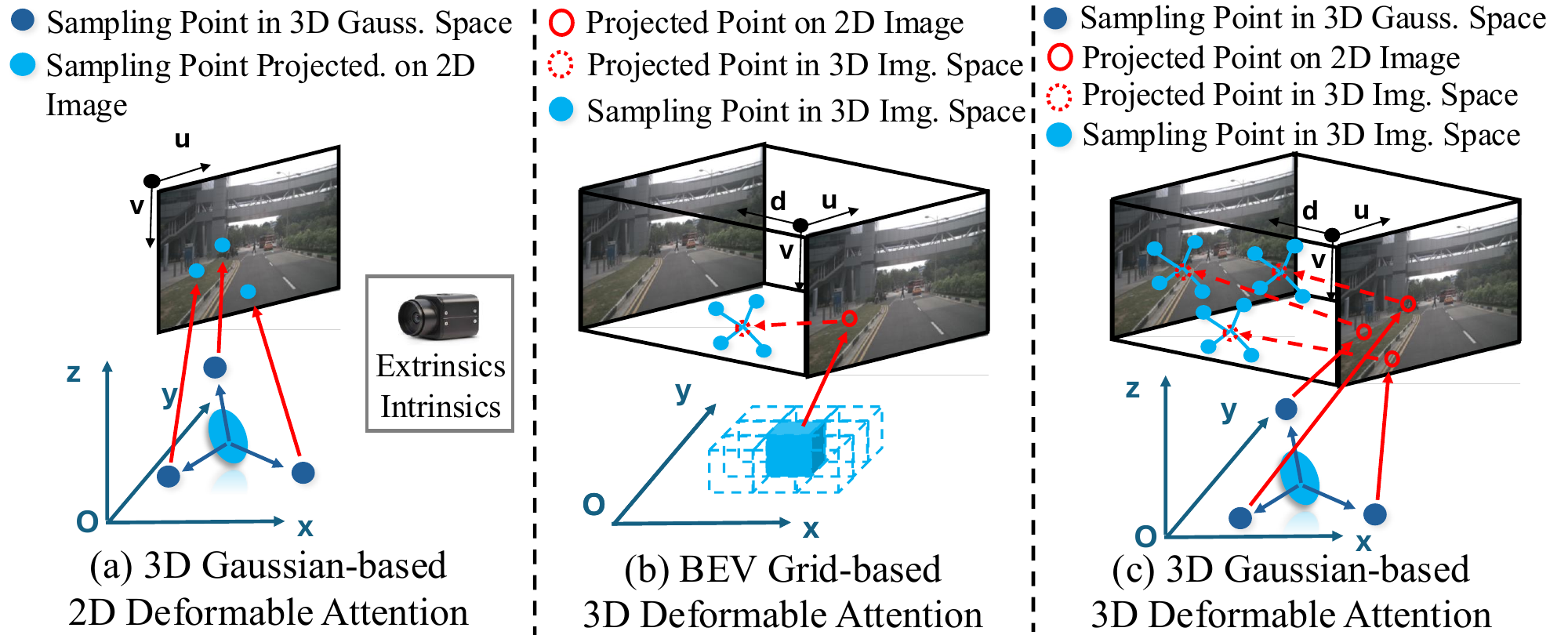}
    \vspace{-15pt}
    \caption{Comparison of different feature sampling methods.}
    \vspace{-5pt}
    \label{fig:sampling_comparison}
\end{figure}

Lift, Splat, Shoot (LSS)~\cite{lss} and 2D attention-based methods~\cite{deformabledetr} are widely adopted for feature lifting which transform multi-view 2D images into a 3D space to obtain lifted features. However, LSS suffers from excessive computational costs, hindering its application to multi-scale feature maps that are important for recognizing objects of various sizes. 
GaussianFormer~\cite{gaussianformer} utilizes a 2D deformable attention (\cref{fig:sampling_comparison}(a)) to extract visual information from 2D images. Despite its efficiency, it suffers from the depth ambiguity problem. As multiple 3D reference points from different Gaussians can be projected to the same 2D position with similar sampling points in the 2D view, this leads to ineffective aggregation, i.e., aggregating the same 2D features for different 3D Gaussian queries. 
The underlying reason for this is the lack of accurate depth information during the feature lifting and aggregating.
A 3D deformable attention operator, namely DFA3D~\cite{dfa3d}, is designed to mitigate the depth ambiguity problem by first expanding 2D feature maps into 3D using estimated depth~\cite{depthnet} and then applying an attention mechanism~\cite{transformer} to aggregate features from the expanded 3D feature maps. However, the operator is originally designed for BEV-based 3D object detection (\cref{fig:sampling_comparison}(b)), and relies on DepthNet~\cite{depthnet} to estimate monocular depth. Inspired by DFA3D~\cite{dfa3d}, we propose a LiDAR-guided 3D deformable attention mechanism for Gaussian-based semantic occupancy prediction, as illustrated in the dashed orange box in \cref{fig:overall}. We first form a unified LiDAR-camera 3D feature space $\mathbf{F}^{{\rm 3D}}$ by conducting outer product between the multi-scale depth maps $\mathbf{F}^{\rm d}$, generated from the LiDAR point cloud, and the multi-scale camera feature maps \mbox{$\mathbf{F}^{\rm c} : \mathbf{F}^{\rm 3D} = \mathbf{F}^{\rm d} \otimes \mathbf{F}^{\rm c}$}.
For feature sampling, we design a two-stage key point sampling method (\cref{fig:sampling_comparison}(c)) to aggregate sufficient informative features for updating Gaussian queries. First, we sample a group of 3D reference points \mbox{$\mathcal{R}_G=\{\mathbf{m}_i=\mathbf{m}+\Delta\mathbf{m}_i | i=1,...,N_{R_1}\}$} for each Gaussian $\mathbf{G}$ by shifting its mean $\mathbf{m}$ with learned offsets $\Delta \mathbf{m}$.
Then we project these 3D reference points into the fusion feature space $\mathbf{F}^{\rm 3D}$ with extrinsics $\mathcal{T}$ and intrinsics $\mathcal{K}$, where each projected reference point is positioned at \mbox{$\mathbf{\bar m}_i = (u_i, v_i, d_i)$}.
After projection, we further generate learnable sampling offsets \mbox{$\Delta \mathbf{\bar m}_{ij} = (\Delta u_{ij}, \Delta v_{ij}, \Delta d_{ij})$} for each projected reference point $\mathbf{\bar m}_i$. The overall sampling points of a given Gaussian $\mathbf{G}$ in the fusion feature space $\mathbf{F}^{\rm 3D}$ can be formulated as:
\begin{equation}
    \mathcal{\bar R}_G=\{\mathbf{\bar m}_{ij} = \mathbf{\bar m}_i + \Delta\mathbf{\bar m}_{ij}| i=1,...,N_{R_1}, j=1,...,N_{R_2}\},
\end{equation}
where $N_{R_1}$ and $N_{R_2}$ denote the number of sampling points for each Gaussian and for each projected 3D reference point.
Finally, the Gaussian query $\mathbf{Q}$ is updated with the weighted sum of aggregated LiDAR-camera fusion features $\Delta \mathbf{Q}$:
\begin{equation}
    \Delta \mathbf{Q} = \frac{1}{N_c}\sum_{c=1}^{N_c}\sum_{i=1}^{N_{R_1}}\sum_{j=1}^{N_{R_2}}{\rm DFA}(\mathbf{Q}, \boldsymbol{\pi}_c(\mathbf{\bar m}_{ij}; \mathcal{T},\mathcal{K}), \mathbf{F}_c^{\rm 3D}),
\end{equation}
where $\rm DFA(\cdot)$ and $\boldsymbol{\pi}_c(\cdot)$ represent the 3D deformable attention and the transformation from the Gaussian frame to $\mathbf{F}_c^{{\rm 3D}}$ frame generated from camera view $c$, respectively. 
After acquiring sufficient geometric and semantic information through sparse convolution and 3D deformable attention, the Gaussian query $\mathbf{Q}$ is passed to a multi-layer perceptron, and decoded to refine the Gaussian property $\mathbf{G}$. We iteratively optimize the properties with 4 blocks of sparse convolution, 3D deformable attention, and refinement modules.

\definecolor{nbarrier}{RGB}{255, 120, 50}
\definecolor{nbicycle}{RGB}{255, 192, 203}
\definecolor{nbus}{RGB}{255, 255, 0}
\definecolor{ncar}{RGB}{0, 150, 245}
\definecolor{nconstruct}{RGB}{0, 255, 255}
\definecolor{nmotor}{RGB}{200, 180, 0}
\definecolor{npedestrian}{RGB}{255, 0, 0}
\definecolor{ntraffic}{RGB}{255, 240, 150}
\definecolor{ntrailer}{RGB}{135, 60, 0}
\definecolor{ntruck}{RGB}{160, 32, 240}
\definecolor{ndriveable}{RGB}{255, 0, 255}
\definecolor{nother}{RGB}{139, 137, 137}
\definecolor{nsidewalk}{RGB}{75, 0, 75}
\definecolor{nterrain}{RGB}{150, 240, 80}
\definecolor{nmanmade}{RGB}{213, 213, 213}
\definecolor{nvegetation}{RGB}{0, 175, 0}\definecolor{tan}{rgb}{0.82, 0.71, 0.55}

\begin{table*}[!h]

  \caption{3D semantic occupancy prediction results on nuScenes-SurroundOcc~\cite{surroundocc} validation set.}
  \vspace{-5pt}
    \centering
  \resizebox{0.95\linewidth}{!}{
    \begin{tabular}{c|c|cc|cccccccccccccccc}
      \toprule
      Method & Modality & IoU $\uparrow$ & mIoU $\uparrow$ & 
      \rotatebox{90}{barrier} & 
      \rotatebox{90}{bicycle} & 
      \rotatebox{90}{bus} & 
      \rotatebox{90}{car} & 
      \rotatebox{90}{const. veh.} & 
      \rotatebox{90}{motorcycle} & 
      \rotatebox{90}{pedestrian} & 
      \rotatebox{90}{traffic cone} & 
      \rotatebox{90}{trailer} & 
      \rotatebox{90}{truck} & 
      \rotatebox{90}{drive. surf.} & 
      \rotatebox{90}{other flat} & 
      \rotatebox{90}{sidewalk} & 
      \rotatebox{90}{terrain} & 
      \rotatebox{90}{manmade} & 
      \rotatebox{90}{vegetation} \\
      & & & & 
      \tikz \draw[fill=nbarrier,draw=nbarrier] (0,0) rectangle (0.2,0.2); & 
      \tikz \draw[fill=nbicycle,draw=nbicycle] (0,0) rectangle (0.2,0.2); & 
      \tikz \draw[fill=nbus,draw=nbus] (0,0) rectangle (0.2,0.2); & 
      \tikz \draw[fill=ncar,draw=ncar] (0,0) rectangle (0.2,0.2); & 
      \tikz \draw[fill=nconstruct,draw=nconstruct] (0,0) rectangle (0.2,0.2); & 
      \tikz \draw[fill=nmotor,draw=nmotor] (0,0) rectangle (0.2,0.2); & 
      \tikz \draw[fill=npedestrian,draw=npedestrian] (0,0) rectangle (0.2,0.2); & 
      \tikz \draw[fill=ntraffic,draw=ntraffic] (0,0) rectangle (0.2,0.2); & 
      \tikz \draw[fill=ntrailer,draw=ntrailer] (0,0) rectangle (0.2,0.2); & 
      \tikz \draw[fill=ntruck,draw=ntruck] (0,0) rectangle (0.2,0.2); & 
      \tikz \draw[fill=ndriveable,draw=ndriveable] (0,0) rectangle (0.2,0.2); & 
      \tikz \draw[fill=nother,draw=nother] (0,0) rectangle (0.2,0.2); & 
      \tikz \draw[fill=nsidewalk,draw=nsidewalk] (0,0) rectangle (0.2,0.2); & 
      \tikz \draw[fill=nterrain,draw=nterrain] (0,0) rectangle (0.2,0.2); & 
      \tikz \draw[fill=nmanmade,draw=nmanmade] (0,0) rectangle (0.2,0.2); & 
      \tikz \draw[fill=nvegetation,draw=nvegetation] (0,0) rectangle (0.2,0.2); \\
      \midrule
      MonoScene~\cite{monoscene} & C  & 24.0 & 7.3 & 4.0 & 0.4 & 8.0 & 8.0 & 2.9 & 0.3 & 1.2 & 0.7 & 4.0 & 4.4 & 27.7 & 5.2 & 15.1 & 11.3 & 9.0 & 14.9\\
      BEVFormer~\cite{bevformer} & C & 30.5& 16.8& 14.2& 6.6& 23.5& 28.3& 8.7& 10.8& 6.6& 4.1& 11.2& 17.8& 37.3& 18.0& 22.9& 22.2& 13.8& 22.2\\
      TPVFormer~\cite{tpvformer} & C & 30.9& 17.1& 16.0& 5.3& 23.9& 27.3& 9.8& 8.7& 7.1& 5.2& 11.0& 19.2& 38.9& 21.3& 24.3& 23.2& 11.7& 20.8\\
      OccFormer~\cite{occformer} & C & 31.4& 19.0& 18.7& 10.4& 23.9& 30.3& 10.3& 14.2& 13.6& 10.1& 12.5& 20.8& 38.8& 19.8& 24.2& 22.2& 13.5& 21.4\\
      SurroundOcc~\cite{surroundocc} & C & 31.5& 20.3& 20.6& 11.7& 28.1& 30.9& 10.7& 15.1& 14.1& 12.1& 14.4& 22.3& 37.3& 23.7& 24.5& 22.8& 14.9& 21.9\\
      C-CONet~\cite{openoccupancy} & C & 26.1& 18.4& 18.6& 10.0& 26.4& 27.4& 8.6& 15.7& 13.3& 9.7& 10.9& 20.2& 33.0& 20.7& 21.4& 21.8& 14.7& 21.3\\
      FB-Occ~\cite{fbocc} & C & 31.5& 19.6& 20.6& 11.3& 26.9& 29.8& 10.4& 13.6& 13.7& 11.4& 11.5& 20.6& 38.2& 21.5& 24.6& 22.7& 14.8& 21.6\\
      GaussianFormer~\cite{gaussianformer} & C & 29.8& 19.1& 19.5& 11.3& 26.1& 29.8& 10.5& 13.8& 12.6& 8.7& 12.7& 21.6& 39.6& 23.3& 24.5& 23.0& 9.6& 19.1\\
      GaussianFormer-2~\cite{gaussianformer2} & C & 31.7& 20.8& 21.4& 13.4& 28.5& 30.8& 10.9& 15.8& 13.6& 10.5& 14.0& 22.9& 40.6& 24.4& 26.1& 24.3& 13.8& 22.0 \\ 
      \midrule
      LMSCNet~\cite{lmscnet} & L & 36.6& 14.9& 13.1& 4.5& 14.7& 22.1& 12.6& 4.2& 7.2& 7.1& 12.2& 11.5& 26.3& 14.3& 21.1& 15.2& 18.5& 34.2\\
      L-CONet~\cite{openoccupancy} & L & 39.4& 17.7& 19.2& 4.0& 15.1& 26.9& 6.2& 3.8& 6.8& 6.0& 14.1& 13.1& 39.7& 19.1& 24.0& 23.9& 25.1& 35.7\\
      \midrule
      M-CONet~\cite{openoccupancy} & L+C & 39.2& 24.7& 24.8& 13.0& 31.6& 34.8& 14.6& 18.0& 20.0& 14.7& 20.0& 26.6& 39.2& 22.8& 26.1& 26.0& 26.0& 37.1 \\
      Co-Occ~\cite{coocc} & L+C & \underline{41.1}& \textbf{27.1}& 28.1& 16.1& 34.0& 37.2& 17.0& 21.6& 20.8& 15.9& 21.9& 28.7& 42.3& 25.4& 29.1& 28.6& 28.2& 38.0 \\
      \rowcolor{black!10}
      \textbf{Ours} & L+C & \textbf{43.3}& \textbf{27.1} & 26.9& 15.8& 32.7& 36.1& 18.6& 21.7& 24.1& 13.0& 21.3& 29.0& 40.6& 23.7& 27.3& 28.2& 32.6& 42.3\\
      \bottomrule
    \end{tabular}
  }
  \vspace{-5pt}
  \label{tab:surroundocc}
\end{table*}

\begin{table*}[!h]
  \caption{3D semantic occupancy prediction results on nuScenes-Occ3D~\cite{tian2024occ3d} validation set. * denotes training with camera visibility mask. (xf) denotes the number of history image frames used for temporal fusion.}
  \vspace{-5pt}
    \centering
  \resizebox{0.95\linewidth}{!}{
    \begin{tabular}{c|c|c|ccccccccccccccccc}
      \toprule
      Method & Modality  & mIoU $\uparrow$ & 
      \rotatebox{90}{others} &
      \rotatebox{90}{barrier} & 
      \rotatebox{90}{bicycle} & 
      \rotatebox{90}{bus} & 
      \rotatebox{90}{car} & 
      \rotatebox{90}{const. veh.} & 
      \rotatebox{90}{motorcycle} & 
      \rotatebox{90}{pedestrian} & 
      \rotatebox{90}{traffic cone} & 
      \rotatebox{90}{trailer} & 
      \rotatebox{90}{truck} & 
      \rotatebox{90}{drive. surf.} & 
      \rotatebox{90}{other flat} & 
      \rotatebox{90}{sidewalk} & 
      \rotatebox{90}{terrain} & 
      \rotatebox{90}{manmade} & 
      \rotatebox{90}{vegetation} \\
      &  & & 
    \tikz \draw[fill=black,draw=black] (0,0) rectangle (0.2,0.2); &      
      \tikz \draw[fill=nbarrier,draw=nbarrier] (0,0) rectangle (0.2,0.2); & 
      \tikz \draw[fill=nbicycle,draw=nbicycle] (0,0) rectangle (0.2,0.2); & 
      \tikz \draw[fill=nbus,draw=nbus] (0,0) rectangle (0.2,0.2); & 
      \tikz \draw[fill=ncar,draw=ncar] (0,0) rectangle (0.2,0.2); & 
      \tikz \draw[fill=nconstruct,draw=nconstruct] (0,0) rectangle (0.2,0.2); & 
      \tikz \draw[fill=nmotor,draw=nmotor] (0,0) rectangle (0.2,0.2); & 
      \tikz \draw[fill=npedestrian,draw=npedestrian] (0,0) rectangle (0.2,0.2); & 
      \tikz \draw[fill=ntraffic,draw=ntraffic] (0,0) rectangle (0.2,0.2); & 
      \tikz \draw[fill=ntrailer,draw=ntrailer] (0,0) rectangle (0.2,0.2); & 
      \tikz \draw[fill=ntruck,draw=ntruck] (0,0) rectangle (0.2,0.2); & 
      \tikz \draw[fill=ndriveable,draw=ndriveable] (0,0) rectangle (0.2,0.2); & 
      \tikz \draw[fill=nother,draw=nother] (0,0) rectangle (0.2,0.2); & 
      \tikz \draw[fill=nsidewalk,draw=nsidewalk] (0,0) rectangle (0.2,0.2); & 
      \tikz \draw[fill=nterrain,draw=nterrain] (0,0) rectangle (0.2,0.2); & 
      \tikz \draw[fill=nmanmade,draw=nmanmade] (0,0) rectangle (0.2,0.2); & 
      \tikz \draw[fill=nvegetation,draw=nvegetation] (0,0) rectangle (0.2,0.2); \\
      \midrule
      MonoScene~\cite{monoscene} & C & 6.1& 1.8& 7.2& 4.3& 4.9& 9.4& 5.7& 4.0& 3.0& 5.9& 4.5& 7.2& 14.9& 6.3& 7.9& 7.4& 1.0& 7.7 \\
      BEVFormer~\cite{bevformer} & C & 23.7& 5.0& 38.8& 10.0& 34.4& 41.1& 13.2& 16.5& 18.2& 17.8& 18.7& 27.7& 49.0& 27.7& 29.1& 25.4& 15.4& 14.5\\
      TPVFormer~\cite{tpvformer} & C & 28.3& 6.7& 39.2& 14.2& 41.5& 47.0& 19.2& 22.6& 17.9& 14.5& 30.2& 35.5& 56.2& 33.7& 35.7& 31.6& 20.0& 16.1\\
      CTF-Occ~\cite{tian2024occ3d} & C & 28.5& 8.1& 39.3& 20.6& 38.3& 42.2& 16.9& 24.5& 22.7& 21.1& 23.0& 31.1& 53.3& 33.8& 38.0& 33.2& 20.8& 18.0\\
      RenderOcc~\cite{renderocc} & C & 26.1& 4.8& 31.7& 10.7& 27.7& 26.5& 13.9& 18.2& 17.7& 17.8& 21.2& 23.3& 63.2& 36.4& 46.2& 44.3& 19.6& 20.7\\
      GaussianFormer*~\cite{gaussianformer} & C & 35.5& 8.8 & 40.9 & 23.3 & 42.9 & 49.7 & 19.2 & 24.8 & 24.4 & 22.5 & 29.4 & 35.3 & 79.0 & 36.9 & 46.6 & 48.2 & 38.8 & 33.1 \\
      COTR* (2f)~\cite{cotr} & C & 44.5& 13.3& 52.1& 32.0& 46.0& 55.6& 32.6& 32.8& 30.4& 34.1& 37.7& 41.8& 84.5& 46.2& 57.6& 60.7& 52.0& 46.3\\
      PanoOcc* (4f)~\cite{panoocc} & C & 42.1& 11.7& 50.5& 29.6& 49.4& 55.5& 23.3& 33.3& 30.6& 31.0& 34.4& 42.6& 83.3& 44.2& 54.4& 56.0& 45.9& 40.4\\
      FB-Occ* (16f)~\cite{fbocc} & C & 42.1 & 14.3& 49.7& 30.0& 46.6& 51.5& 29.3& 29.1& 29.4& 30.5& 35.0& 39.4& 83.1& 47.2& 55.6& 59.9& 44.9& 39.6\\
      \midrule
      OccFusion*~\cite{occfusiondepthfree} & L+C & \textbf{48.7} & 12.4 & 51.8 & 33.0& 54.6& 57.7 & 34.0 & 43.0 & 48.4 & 35.5& 41.2& 48.6& 83.0& 44.7& 57.1& 60.0& 62.5& 61.3 \\
      \rowcolor{black!10}
      \textbf{Ours*} & L+C & \underline{46.4} & 9.8 & 50.0 & 31.3 & 54.0 & 59.4 & 28.1 & 36.2 & 46.2 & 26.7 & 40.2 &  49.7 & 79.1 & 37.3 & 49.0 & 55.0 & 69.1 & 67.6\\
      \bottomrule
    \end{tabular}
  }
  \vspace{-10pt}
  \label{tab:occ3d}
\end{table*}

\section{Experiments}

\subsection{Datasets}

\textbf{NuScenes}~\cite{nuscenes} dataset provides 1000 sequences of driving scenes collected with 6 surrounding cameras, 1 LiDAR and 5 radars. Each sequence lasts 20 seconds and is annotated at a frequency of 2$\rm Hz$.
\textbf{SurroundOcc}~\cite{surroundocc} and \textbf{Occ3D}~\cite{tian2024occ3d} both provide semantic occupancy annotation for nuScenes dataset, each including 700 and 150 scenes for training and validation respectively, for 18 classes (i.e., 16 semantics, 1 noise class and 1 empty). Differently, SurroundOcc partitions each scene within the range of $[-50\rm{m}, 50\rm{m}] \times [-50\rm{m}, 50\rm{m}] \times [-5\rm{m}, 3\rm{m}]$ into voxels with a resolution of $0.5\mathrm{m}$, whereas Occ3D divides a scene within $[-40\rm{m}, 40\rm{m}] \times [-40\rm{m}, 40\rm{m}] \times [-1\rm{m}, 5.4\rm{m}]$ into voxels with a resolution of $0.4\mathrm{m}$. A camera visibility mask is also provided in Occ3D.

\textbf{RELLIS-3D}~\cite{rellis3d} dataset is a multi-modal off-road driving dataset collected by a Clearpath Warthog robot containing RGB images, LiDAR point clouds, stereo images, GPS and IMU data. \textbf{WildOcc}~\cite{wildocc} provides the first off-road occupancy annotation on the RELLIS-3D, which are split into 7399/1249/1399 frames for training, validation and testing respectively. The annotation is in the range of $[-20\rm{m}, 0\rm{m}] \times [-10\rm{m}, 10\rm{m}] \times [-2\rm{m}, 6\rm{m}]$, where each voxel has a resolution of $0.2\mathrm{m}$ and labeled as one of 9 classes (7 semantics, 1 other class and 1 empty). WildOcc~\cite{wildocc} is used to evaluate the performance of our model in complex off-road environments and with a monocular-LiDAR sensor configuration.


\subsection{Implementation and Evaluation Details}
For camera branch, we set the resolution of input images as $900 \times 1600$ for nuScenes~\cite{nuscenes} and $1200\times1920$ for RELLIS-3D~\cite{rellis3d}. We utilize the ResNet101-DCN~\cite{resnet} checkpoint pretrained from FCOS3D~\cite{fcos3d} as the backbone and FPN~\cite{fpn} as the neck. For LiDAR branch, we aggregate and voxelize previous 10 sweeps of point clouds, and obtain the mean features through a voxel feature encoder~\cite{voxelnet}. The LiDAR depth map is generated and saved before training following~\cite{dfa3d, bevdepth}. The number of Gaussians is set to 25,600 in our main experiments. We employ these Gaussians to only model the occupied space, and leave the empty space to one fixed large Gaussian to improve efficiency~\cite{gaussianformer2}.
We train our model with an AdamW optimizer with a weight decay of 0.01. The learning rates are set to $1 \times 10^{-4}$ for nuScenes and $3 \times 10^{-4}$ for RELLIS-3D, and decay with a cosine annealing schedule. 
Our model is trained for 24 epochs with a batch size of 8 on nuScenes and 20 epochs with a batch size of 4 on RELLIS-3D on Nvidia A40 GPUs. We use Intersection-over-Union (IoU) and mean Intersection-over-Union (mIoU) for evaluation metrics following MonoScene~\cite{monoscene}. 

\definecolor{dirt}{rgb}{0.43, 0.08, 0.54}
\definecolor{grass}{rgb}{0, 0.5, 0}
\definecolor{tree}{rgb}{0, 1, 0} 
\definecolor{bush}{rgb}{1, 0.6, 0.8}
\definecolor{Barrier}{rgb}{0.16, 0.5, 1}
\definecolor{Puddle}{rgb}{0.52, 1, 0.94}
\definecolor{Mud}{rgb}{0.39, 0.26, 0.13}

\begin{table}[!h]

  \caption{3D semantic occupancy prediction results on RELLIS3D-WildOcc~\cite{wildocc} dataset.}
  \vspace{-5pt}
    \centering
  \resizebox{\linewidth}{!}{
    \begin{tabular}{c|c|cc|ccccccc}
      \toprule
      Method & Mod. & IoU $\uparrow$ & mIoU $\uparrow$ & 
      \rotatebox{90}{{\centering Grass}} &  
      \rotatebox{90}{{\centering Tree}} &  
      \rotatebox{90}{{\centering Bush}} &
      \rotatebox{90}{{\centering Puddle}} &  
      \rotatebox{90}{{\centering Mud}} &  
      \rotatebox{90}{{\centering Barrie}} &    
      \rotatebox{90}{{\centering Rubble}}  \\
      & & & & 
      \tikz \draw[fill=grass,draw=grass] (0,0) rectangle (0.2,0.2); & 
      \tikz \draw[fill=tree,draw=tree] (0,0) rectangle (0.2,0.2); & 
      \tikz \draw[fill=bush,draw=bush] (0,0) rectangle (0.2,0.2); & 
      \tikz \draw[fill=Puddle,draw=Puddle] (0,0) rectangle (0.2,0.2); & 
      \tikz \draw[fill=Mud,draw=Mud] (0,0) rectangle (0.2,0.2); & 
      \tikz \draw[fill=Barrier,draw=Barrier] (0,0) rectangle (0.2,0.2); & 
      \tikz \draw[fill=dirt,draw=dirt] (0,0) rectangle (0.2,0.2); \\
      \midrule
      \multicolumn{1}{c|}{\emph{\textbf{Test}} Set} & \multicolumn{3}{c|}{Class Percentage \%} &  41.052 & 36.094 & 17.621 & 0.512 & 0.774 & 0.001 & 0.001\\
      \midrule
      C-OFFOcc (4f) ~\cite{wildocc} & C & 29.7& 11.2& 24.6& 23.8& 22.1& 0.6& 3.5& 0.6 & 3.2\\
      GaussianFormer~\cite{gaussianformer} & C & 19.5 & 6.3 & 21.8 & 12.1 & 5.2 & 2.7 & 2.3 & 0.0 & 0.0\\
      \midrule
      M-OFFOcc~\cite{wildocc} & L+C & - & 12.9 & - & - & - & - & - & - & -\\
      M-OFFOcc (4f)~\cite{wildocc} & L+C & 32.8 & \textbf{14.8} & 28.6 & 33.4 & 27.5 & 0.9& 6.8 & 1.7 & 4.6\\
      \rowcolor{black!10}
      \textbf{Ours} & L+C & \textbf{33.9} & 13.1 & 24.0 & 45.4 & 12.9 & 6.6 & 2.8 & 0.0 & 0.0 \\
      \midrule
      \multicolumn{1}{c|}{\emph{\textbf{Validation}} Set} & \multicolumn{3}{c|}{Class Percentage \%} & 31.739 & 42.210 & 18.497 & 0.105 & 0.842 & 2.218 & 3.836\\
      \midrule
      GaussianFormer~\cite{gaussianformer} & C & 23.0 & 8.2 & 19.4 & 24.4 & 5.2 & 0.0 & 4.4 & 0.0 & 4.0\\
      \midrule
      \rowcolor{black!10}
      \textbf{Ours} & L+C & \textbf{29.5} & \textbf{13.1} & 19.1 & 38.5 & 10.6 & 0.1 & 4.6 & 4.2 & 14.5 \\
      \bottomrule
    \end{tabular}
  }
  \vspace{-5pt}
  \label{tab:wildocc}
\end{table}

\begin{table}[!h]
  \centering
  \caption{Efficiency comparison and ablation on number of Gaussians. Tested on one A40 GPU with one batch during inference.}
  \vspace{-5pt}
  \resizebox{\linewidth}{!}{
    \begin{tabular}{c|c|c|c|cc|cc}
      \toprule
      Method & Mod. & \makecell{Query\\Form} & \makecell{Query\\Number} &  \makecell{Lat. \\ (ms) $\downarrow$} & \makecell{Mem. \\ (GB) $\downarrow$} & IoU $\uparrow$ & mIoU $\uparrow$ \\
      \midrule
      BEVFormer~\cite{bevformer} & C & 2D BEV & 200$\times$200 & 310 & 4.5 & 30.5 & 16.8\\
      TPVFormer~\cite{tpvformer} & C & 3D TPV & 200$\times$(200+16+16) & 320 & 5.1 & 30.9 & 17.1\\
      SurroundOcc~\cite{surroundocc} & C & 3D Voxel & 200$\times$200$\times$16 & 340 & 5.9 & \textbf{31.5} & \textbf{20.3}\\
      \cmidrule(lr){2-8}
      ~ & ~ & ~ & 25600 & \textbf{227} & 4.7 & 28.7 & 16.0 \\
      \multirow{-2}*{GaussianFormer~\cite{gaussianformer}} & \multirow{-2}*{C} & \multirow{-2}*{3D Gaussian} & 144000 & 370 & 6.1 & 29.8 & 19.1\\
      \cmidrule(lr){2-8}
      ~ & ~  & ~ & 6400 & 313 & \textbf{3.0} & 30.4 & 19.9 \\
      ~ & ~ & ~ & 12800 & 323 & \textbf{3.0} & 30.4 & 19.9 \\
      \multirow{-3}*{GaussianFormer-2~\cite{gaussianformer2}} & \multirow{-3}*{C} & \multirow{-3}*{3D Gaussian} & 25600 & 357 & \textbf{3.0} & 31.0 & \textbf{20.3}\\
      \midrule
      ~ & ~ & ~ & 50$\times$50$\times$4 & 532 & 7.6 & 33.3 & 21.2\\
      \multirow{-2}*{M-CONet~\cite{openoccupancy}} & \multirow{-2}*{L+C} & \multirow{-2}*{3D Voxel} & 100$\times$100$\times$8 & 670 & 7.8 & 39.2 & 24.7\\
      \cmidrule(lr){2-8}
      Co-Occ~\cite{coocc} & L+C & 3D Voxel & 100$\times$100$\times$8 & 580 & 11.8 & 41.1 & \textbf{27.1}\\
      \cmidrule(lr){2-8}
      \rowcolor{black!10}
      ~ & ~ & ~ & 6400 & \textbf{415} & \textbf{4.9} & 39.6 & 21.4 \\
      \rowcolor{black!10}
      ~ & ~ & ~ & 12800 & 462 & 5.0 & 41.4 & 24.2 \\
      \rowcolor{black!10}
       \multirow{-3}*{\textbf{Ours}} & \multirow{-3}*{L+C} & \multirow{-3}*{3D Gaussian} & 25600 & 555 & 5.5 & \textbf{43.3} & \textbf{27.1}\\
      \bottomrule
    \end{tabular}
}
  \label{tab:efficiency_and_gaussiannumber}
  \vspace{-15pt}
\end{table}

\subsection{Quantitative Results}

\textbf{3D semantic occupancy prediction performance.} We report the performance of GaussianFormer3D on SurroundOcc~\cite{surroundocc}, Occ3D~\cite{tian2024occ3d} and WildOcc~\cite{wildocc} in \cref{tab:surroundocc}, \cref{tab:occ3d} and \cref{tab:wildocc}, respectively.
For on-road scenarios in \cref{tab:surroundocc} and \cref{tab:occ3d}, our method surpasses GaussianFormer~\cite{gaussianformer} extensively on all classes, leading to overall 13.5 and 8.0 increases on the IoU and mIoU respectively on SurroundOcc~\cite{surroundocc} and 10.9 increase on the mIoU on Occ3D~\cite{tian2024occ3d}. Compared to state-of-the-art LiDAR-camera approaches~\cite{coocc, openoccupancy, occfusiondepthfree}, ours achieves best overall performance while showing superior performance in predicting small objects (e.g., \textit{motorcycle}, \textit{pedestrian}), dynamic vehicles (e.g., \textit{car}, \textit{construction vehicle}, \textit{truck}) and surrounding surfaces (e.g., \textit{manmade}, \textit{vegetation}) which are crucial classes for autonomous driving tasks. This improvement is due to Gaussians' universal approximating ability to model objects with flexible scales and shapes.
For off-road results in \cref{tab:wildocc}, our method with single-frame image input surpasses M-OFFOcc~\cite{wildocc} using 4 sequential images by 1.1 in IoU and performs on par in mIoU. Moreover, our method outperforms GaussianFormer~\cite{gaussianformer} by 14.4 in IoU and 6.8 in mIoU on the test set, highlighting LiDAR's role in understanding the geometry of complex off-road terrains. Our method excels in predicting regions with large surfaces, such as \textit{grass}, \textit{tree}, and \textit{puddle}, while remaining suboptimal for subtle terrain variations like \textit{mud}. For \textit{barrier} and \textit{rubble}, their low occurrence in the test set (0.001\% of occupied voxels) poses a challenge due to the lack of sufficient features for reliable prediction. 
We further evaluate the model performance under different weather conditions in \cref{tab:weather}. Ours shows a significant performance improvement over the baseline under extreme climate (rainy) and low lighting condition (night).

\textbf{Evaluation of model efficiency.} We evaluate and compare the latency and average memory consumption of ours with other methods during testing in \cref{tab:efficiency_and_gaussiannumber}. Ours achieves multi-modal fusion-based prediction performance while maintaining approximately the same low memory usage as camera-only methods. Compared to Co-Occ~\cite{coocc}, ours saves about 50\% average memory consumption, making it more suitable for running onboard on autonomous vehicles. In addition, our approach employs only 25,600 Gaussians with 28 channels while Co-Occ~\cite{coocc} requires 80,000 queries with 128 channels to achieve similar performance, demonstrating the potential of our method to enable more efficient communication for connected vehicles or multi-robot collaborations.
The latency of our method is higher than that of camera-only pipelines, which is mainly due to the computation overhead introduced by 3D deformable attention. 
Some LiDAR-camera methods~\cite{occfusiondepthfree, wildocc} are not compared due to lack of open-source code.
We also examine the effect of the number of Gaussians on the model performance in~\cref{tab:efficiency_and_gaussiannumber}. As the number of Gaussians increases, both latency and memory consumption rise, while the IoU and mIoU metrics are steadily improved.

\begin{figure*}[!t]
\centering
\includegraphics[height=6.8cm]{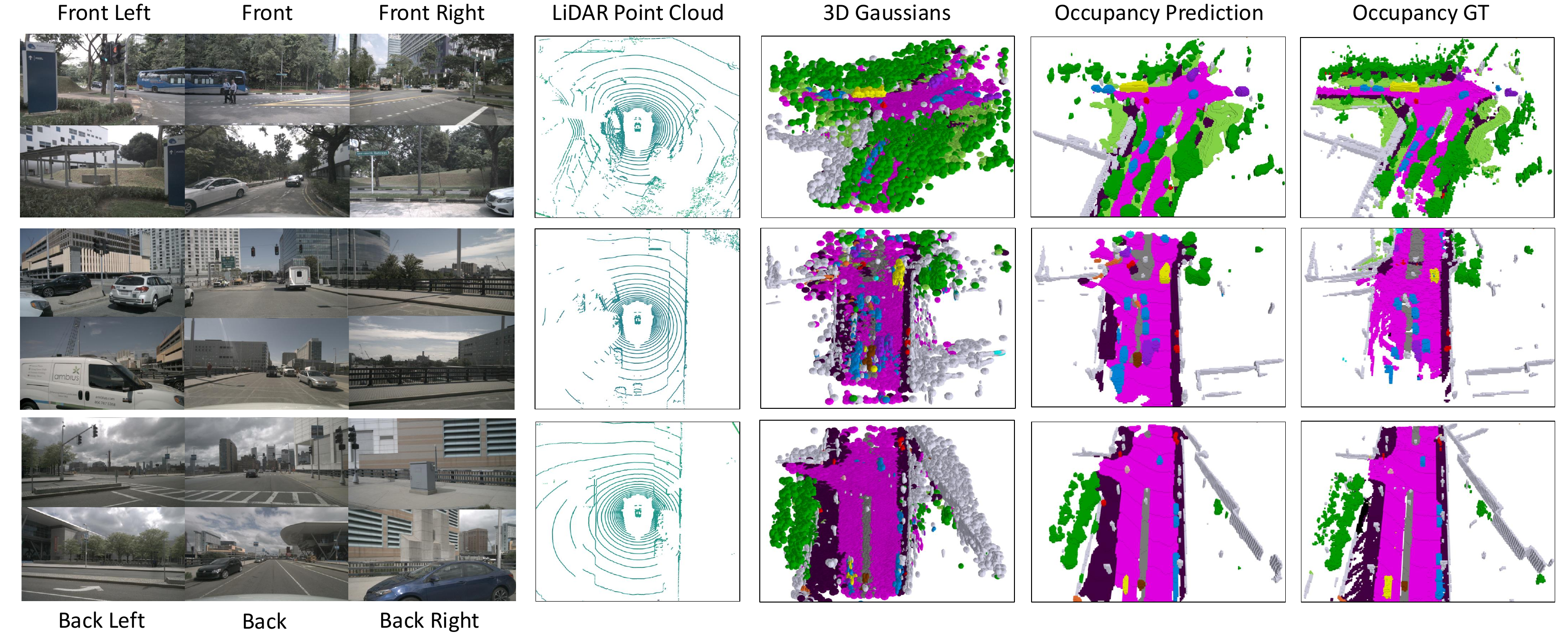}
\vspace{-5pt}
\caption{Qualitative results on the on-road nuScenes-SurroundOcc~\cite{surroundocc} validation set. Our multi-modal Gaussian-based occupancy method can capture both semantics information and geometry structure of the surroundings. Color legend is given in~\cref{tab:surroundocc}.}
\vspace{-5pt}
\label{fig:main_qualitative}
\end{figure*}

\begin{table*}[htbp]
    \centering
    \caption{Performance on nuScenes-SurroundOcc~\cite{surroundocc} validation set under different weather and lighting conditions.}
    \vspace{-5pt}
    \resizebox{\linewidth}{!}{
     \begin{tabular}{c|c|cccc|cccc}
        \toprule
        & & \multicolumn{4}{c|}{IoU$\uparrow$} & \multicolumn{4}{c}{mIoU$\uparrow$}         \\
        \multirow{-2}{*}{Method} & \multirow{-2}{*}{Modality} & \emoji{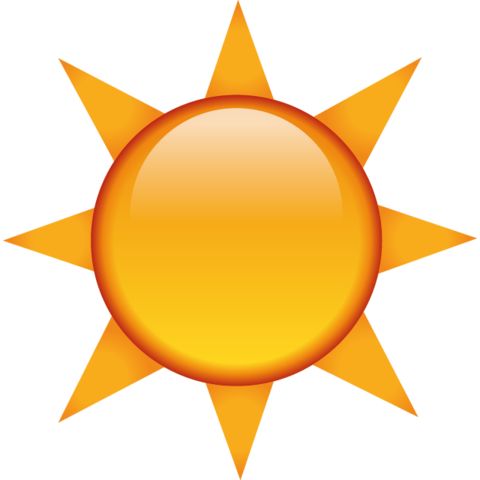} Sunny & \emoji{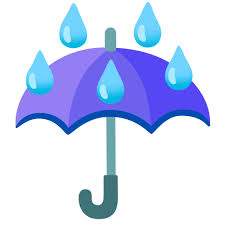} Rainy & \emoji{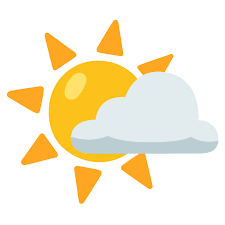} Day & \emoji{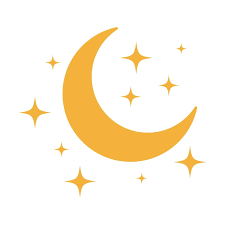} Night & \emoji{figures/sun.jpg} Sunny & \emoji{figures/rainy.jpeg} Rainy & \emoji{figures/day.png} Day & \emoji{figures/night.png} Night\\ 
        \midrule
        GaussianFormer~\cite{gaussianformer} & C & 29.6  & 27.5 & 30.3 & 19.5 & 18.9 & 18.0 & 19.2 & 9.3\\
        \rowcolor{black!10}
        \cellcolor{white} 
        \textbf{GaussianFormer3D} & L+C & \textbf{43.6 (+14.0)} & \textbf{41.6 (\textcolor{red}{+14.1})} & \textbf{43.6 (+13.3)} & \textbf{40.5 (\textcolor{red}{+21.0})} & \textbf{27.3 (+8.4)} & \textbf{25.2 (\textcolor{red}{+7.2})} & \textbf{27.4 (+8.2)} & \textbf{15.5 (\textcolor{red}{+6.2})}\\
        \bottomrule
      \end{tabular}
    }
    \vspace{-3pt}
    \label{tab:weather}
\end{table*}

\begin{table*}[!h]
  \centering
  \caption{Ablation study of proposed modules evaluated on nuScenes-SurroundOcc~\cite{surroundocc} validation set. Voxel-to-Gaussian and LiDAR-Guided 3D Deformable Attention are abbreviated as V2G and DFA respectively.}
  \vspace{-5pt}
  \resizebox{0.95\linewidth}{!}{
    \begin{tabular}{c|cc|cc|cccccccccccccccc}
      \toprule
      Model & V2G & DFA &  IoU $\uparrow$ & mIoU $\uparrow$ &
      \rotatebox{90}{barrier} & 
      \rotatebox{90}{bicycle} & 
      \rotatebox{90}{bus} & 
      \rotatebox{90}{car} & 
      \rotatebox{90}{const. veh.} & 
      \rotatebox{90}{motorcycle} & 
      \rotatebox{90}{pedestrian} & 
      \rotatebox{90}{traffic cone} & 
      \rotatebox{90}{trailer} & 
      \rotatebox{90}{truck} & 
      \rotatebox{90}{drive. surf.} & 
      \rotatebox{90}{other flat} & 
      \rotatebox{90}{sidewalk} & 
      \rotatebox{90}{terrain} & 
      \rotatebox{90}{manmade} & 
      \rotatebox{90}{vegetation} \\
      &  & & & &
      \tikz \draw[fill=nbarrier,draw=nbarrier] (0,0) rectangle (0.2,0.2); & 
      \tikz \draw[fill=nbicycle,draw=nbicycle] (0,0) rectangle (0.2,0.2); & 
      \tikz \draw[fill=nbus,draw=nbus] (0,0) rectangle (0.2,0.2); & 
      \tikz \draw[fill=ncar,draw=ncar] (0,0) rectangle (0.2,0.2); & 
      \tikz \draw[fill=nconstruct,draw=nconstruct] (0,0) rectangle (0.2,0.2); & 
      \tikz \draw[fill=nmotor,draw=nmotor] (0,0) rectangle (0.2,0.2); & 
      \tikz \draw[fill=npedestrian,draw=npedestrian] (0,0) rectangle (0.2,0.2); & 
      \tikz \draw[fill=ntraffic,draw=ntraffic] (0,0) rectangle (0.2,0.2); & 
      \tikz \draw[fill=ntrailer,draw=ntrailer] (0,0) rectangle (0.2,0.2); & 
      \tikz \draw[fill=ntruck,draw=ntruck] (0,0) rectangle (0.2,0.2); & 
      \tikz \draw[fill=ndriveable,draw=ndriveable] (0,0) rectangle (0.2,0.2); & 
      \tikz \draw[fill=nother,draw=nother] (0,0) rectangle (0.2,0.2); & 
      \tikz \draw[fill=nsidewalk,draw=nsidewalk] (0,0) rectangle (0.2,0.2); & 
      \tikz \draw[fill=nterrain,draw=nterrain] (0,0) rectangle (0.2,0.2); & 
      \tikz \draw[fill=nmanmade,draw=nmanmade] (0,0) rectangle (0.2,0.2); & 
      \tikz \draw[fill=nvegetation,draw=nvegetation] (0,0) rectangle (0.2,0.2); \\
      \midrule
      ~ & & & 29.2 & 18.8 & 18.8 & 11.6 & 24.6 & 29.4 & 10.2 & 14.8 & 12.3 & 8.6 & 11.9 & 21.1 & 39.5 & 23.6 & 24.3 & 22.4 & 9.5  & 18.8\\
      \rowcolor{black!5}
      \cellcolor{white}~ & \checkmark & & 40.7 & 25.8 & 25.3 & 17.1 & 30.9 & 35.0 & 17.9 & 21.5 & 23.9 & 14.8 & 20.3 & 27.7 & 37.8 & 20.4 & 24.9 & 25.3 & 30.1 & 39.4\\
      \rowcolor{black!5}
        \cellcolor{white}~ & & \checkmark & 40.7 & 26.4 & 25.3 & 17.4 & 32.4 & 35.7 & 17.8 & 23.9 & 22.1 & 12.0 & 20.5 & 29.1 & 41.8 & 24.6 & 28.1 & 27.7 & 27.5 & 36.6\\
        \rowcolor{black!15}
        \cellcolor{white} 
        \multirow{-4}*{GaussianFormer3D} & \checkmark & \checkmark & 43.3 & 27.1 & 26.9 & 15.8 & 32.7 & 36.1 & 18.6 & 21.7 & 24.1 & 13.0 & 21.3 & 29.0 & 40.6 & 23.7 & 27.3 & 28.2 & 32.6 & 42.3 \\
      \bottomrule
    \end{tabular}
}
  \label{tab:main_ablation}
  \vspace{-5pt}
\end{table*}

\begin{table*}[!h]
    \centering
    \vspace{+3pt}
    \caption{Ablation study of module design choices on the nuScenes-SurroundOcc~\cite{surroundocc} validation set.}
    \vspace{-3pt}
    \scriptsize
    \setlength{\tabcolsep}{3pt}
    \renewcommand{\arraystretch}{0.85}
    \begin{subtable}{0.48\linewidth}
        \centering
        \caption{Ablation study of Gaussian initialization strategies. PM-\\Point denotes probabilistic modeling with point cloud in~\cite{gaussianformer2}.}
        \vspace{-3pt}
      \begin{tabular}{c|ccc|cc}
        \toprule
        Module & Single-Sweep Point & PM-Point & Multi-Sweep Voxel & IoU$\uparrow$ & mIoU$\uparrow$ \\
        \midrule
        ~ & \checkmark &  & & 36.7  & 22.4 \\
        ~ & ~ & \checkmark &  & 34.9  & 21.2 \\
        \rowcolor{black!15}
        \cellcolor{white} 
        \multirow{-3}*{V2G} & &  & \checkmark & 40.7 & 25.8\\
        \bottomrule
      \end{tabular}
      \label{tab:v2g1}
    \end{subtable}
    \vspace{+5pt}
    \begin{subtable}{0.48\linewidth}
        \centering
        \caption{Ablation study of LiDAR voxel size for V2G. The unit of length is $\mathrm{m}$. We set the height of all the voxels as $0.2\mathrm{m}$.}
        \vspace{-3pt}
      \begin{tabular}{c|ccc|cc}
        \toprule
        Module & $0.15 \times 0.15$ & $0.1 \times 0.1$ & $0.075 \times 0.075$ & IoU$\uparrow$ & mIoU$\uparrow$ \\
        \midrule
        ~ & \checkmark &  & & 40.1 & 25.0\\
        ~ & & \checkmark & & 40.6 & 25.2\\
        \rowcolor{black!15}
        \cellcolor{white} 
        \multirow{-3}*{V2G} &  & & \checkmark & 40.7 & 25.8\\
        \bottomrule
      \end{tabular}
      \label{tab:v2g2}
    \end{subtable}
    \vspace{+5pt}
    \begin{subtable}{0.48\linewidth}
        \caption{Ablation study of offset sampling methods for DFA. We run experiments with applying learnable offset sampling before and after projecting Gaussians into the lifted 3D feature space.}
        \vspace{-3pt}
            \centering
      \begin{tabular}{c|cc|cc}
        \toprule
        Module & Sampling Before Projection & Sampling After Projection  & IoU$\uparrow$ & mIoU$\uparrow$ \\
        \midrule
        ~ & \checkmark &  & 37.7 & 24.5\\
        ~ & & \checkmark  &  40.1 & 26.1\\
        \rowcolor{black!15}
        \cellcolor{white} \multirow{-3}*{DFA} & \checkmark &  \checkmark & 40.7 & 26.4\\
        \bottomrule
      \end{tabular}
      \label{tab:dfa1}
    \end{subtable}
    \hspace{2mm}
    \begin{subtable}{0.48\linewidth}
        \centering
        \caption{Ablation study of feature lifting and aggregating methods for DFA. We concatenate LiDAR sparse and dense depth maps with RGB features respectively to conduct 2D deformable attention.}
        \vspace{-3pt}
      \begin{tabular}{c|ccc|cc}
        \toprule
        Module & 2D-Sparse Depth Map & 2D-Dense Depth Map & 3D & IoU$\uparrow$ & mIoU$\uparrow$ \\
        \midrule
        ~ & \checkmark &  & & 36.1 & 22.2 \\
        ~ & & \checkmark &  & 36.6 & 22.1 \\
        \rowcolor{black!15}
        \cellcolor{white} \multirow{-3}*{DFA} & & & \checkmark & 40.7 & 26.4\\
        \bottomrule
      \end{tabular}
      \label{tab:dfa2}
    \end{subtable}
\label{tab:module_ablation}
\vspace{-20pt}
\end{table*}


\begin{figure*}[!h]
    \centering
    \includegraphics[height=4.5cm]{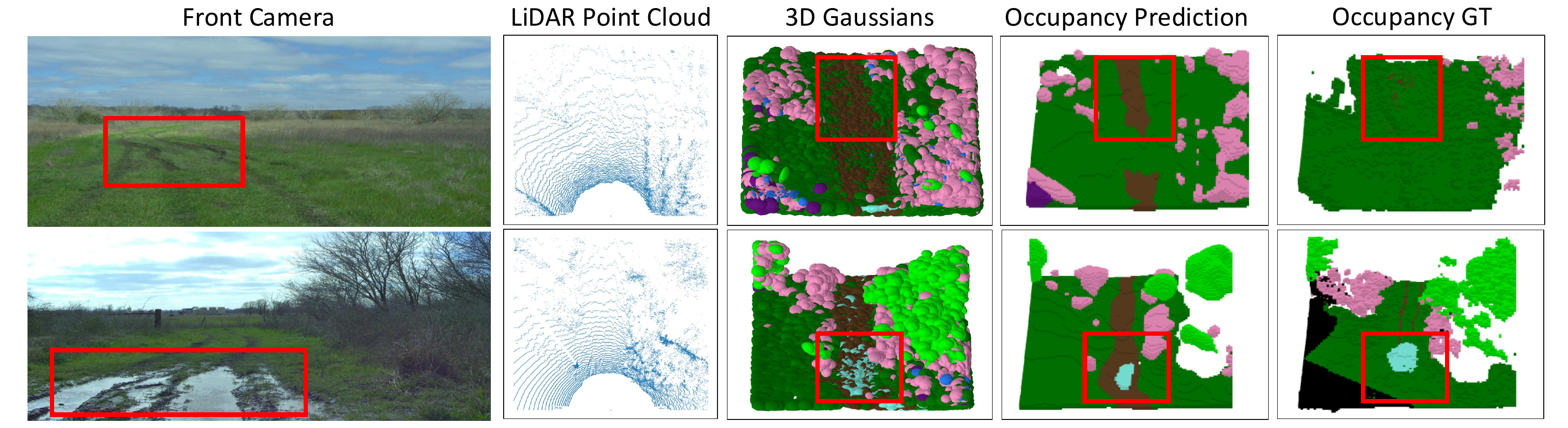}
    \vspace{-5pt}
    \caption{Qualitative results on the off-road RELLIS3D-WildOcc~\cite{wildocc} test set. Our method can outperform the GT (first row) at some regions and predict classes such as \textit{puddle} that are vital for off-road autonomous driving (second row). Color legend is given in~\cref{tab:wildocc}.}
    \vspace{-10pt}
    \label{fig:wildocc}
    \end{figure*}

\begin{figure*}[!h]
\centering
\includegraphics[height=4.9cm]{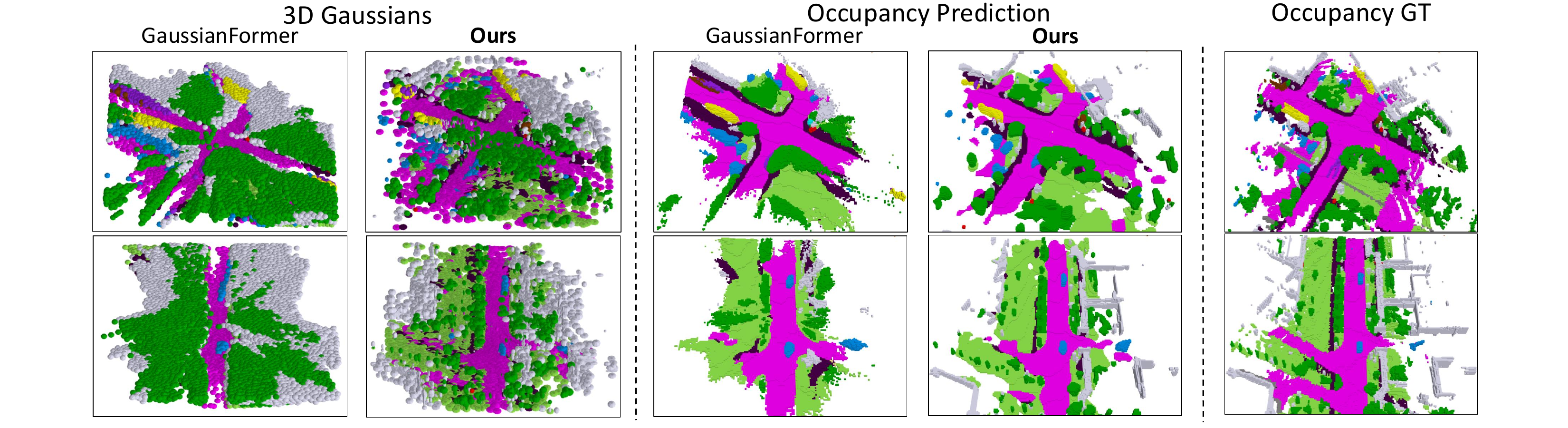}
\vspace{-18pt}
\captionof{figure}{Visualization comparison with GaussianFormer~\cite{gaussianformer} on nuScenes-SurroundOcc~\cite{surroundocc}. By incorporating LiDAR, our method can obtain Gaussians with more adaptive scales and shapes, resulting in more accurate semantic predictions and delicate geometry details.}
\vspace{-15pt}
\label{fig:gaussian_comparison}
\end{figure*}


\begin{figure}[!h]
    \centering
    \includegraphics[width=\linewidth]{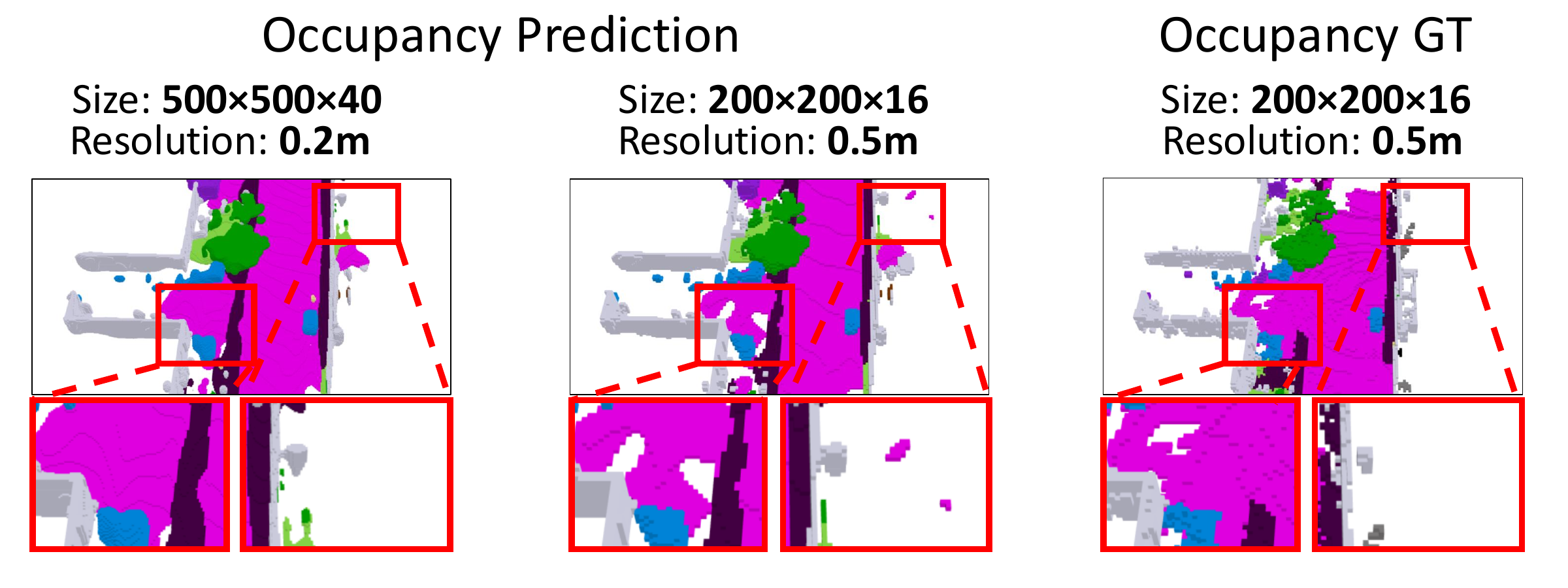}
    \vspace{-18pt}
    \caption{Multi-resolution occupancy prediction of 3D Gaussians.}
    \vspace{-15pt}
    \label{fig:multiresolution}
\end{figure}

\subsection{Ablation Study}
We conduct extensive ablation experiments to validate our design choices. The main ablation study is conducted in \cref{tab:main_ablation}. We observe that both the proposed voxel-to-Gaussian initialization and the LiDAR-guided 3D deformable attention modules contribute to the superior performance of our method. The voxel-to-Gaussian initialization significantly improves the model's ability to detect both small objects (e.g., \textit{pedestrian}, \textit{traffic cone}) and large surfaces (e.g., \textit{manmade}, \textit{vegetation}). This validates the effectiveness of multi-sweep LiDAR scans in providing Gaussians with accurate geometric information of occupied space. We also notice that LiDAR-guided 3D deformable attention mechanism enhances the model's prediction ability on dynamic vehicles (e.g., \textit{bicycle, bus, car, motorcycle, trailer, truck}) and near-road surfaces (e.g., \textit{drivable surface, flatten area, sidewalk, terrain}) where objects detected by LiDAR points are visible to surrounding cameras.
In these regions, the LiDAR points and corresponding image pixels are associated in the lifted 3D feature space, enabling the model to retrieve aggregated fusion features of on-road and near-road objects.

\textbf{Voxel-to-Gaussian Initialization.} We first compare different levels of LiDAR features used for initializing Gaussian properties in~\cref{tab:v2g1}. The improvement achieved with the multi-sweep voxel feature is significantly greater than that of the single-sweep point feature and the point cloud probabilistic modeling strategy used in GaussianFormer-2~\cite{gaussianformer2}, which validates the effectiveness of our proposed module. We further conduct an ablation study on the size of LiDAR voxel in initialization in~\cref{tab:v2g2}. As the voxel size decreases, the model performance slightly improves. We choose $0.075\rm{m} \times 0.075\rm{m} \times 0.2\rm{m}$ as the final size.

\textbf{LiDAR-Guided 3D Deformable Attention.} We first study the effect of the two-stage offset sampling strategy in \cref{tab:dfa1}. We observe that applying learnable offset sampling both before and after projection achieves higher performance than single-stage sampling, which validates our two-stage sampling method can aggregate sufficient informative features for refining Gaussians. We also compare different feature aggregating methods in \cref{tab:dfa2}, including 3D deformable attention, 2D deformable attention with concatenated LiDAR sparse depth map and with completed dense depth map~\cite{fasetdepthcompletion}. The results validate our final choice.

\subsection{Qualitative Results}
We visualize 3D Gaussians and occupancy to qualitatively verify the effectiveness of our method for on-road scenes in~\cref{fig:main_qualitative}. Our method can accurately predict both semantics and fine-grained geometry of the surrounding environments. In some cases, it even outperforms the GT by correctly completing occupancy in regions that lack semantic annotations. 
Qualitative results of our method on off-road scenes are given in~\cref{fig:wildocc}. Our method is able to predict semantic occupancy for classes like \textit{mud} and \textit{puddle}, which are essential for achieving safe and effective off-road autonomous driving.
We further compare our approach with GaussianFormer~\cite{gaussianformer} in~\cref{fig:gaussian_comparison}. The Gaussians in our method are more adaptive in scales and shapes, precisely appearing in the occupied regions of objects in both long-range and short-range areas, aided by the LiDAR sensor.
Additionally, compared to voxel-based discretized approaches that train and predict at a fixed resolution, our method can predict multi-resolution semantic occupancy without additional training cost, attributed to the continuous property of Gaussians. This property enables more accurate and smoother prediction for certain areas when inferred at a higher resolution, as demonstrated in~\cref{fig:multiresolution}.

\section{Conclusion}
In this paper, we proposed GaussianFormer3D, a novel multi-modal semantic occupancy prediction framework that builds on 3D Gaussian scene representation. We introduced a voxel-to-Gaussian initialization strategy to endow 3D Gaussians with accurate geometry priors from LiDAR data. We also designed a LiDAR-guided 3D deformable attention mechanism to refine 3D Gaussians with LiDAR-camera fusion feature. Extensive experiments show its effectiveness in achieving accurate and fine-grained semantic occupancy prediction. 
However, our model is limited to fully-supervised manner, which requires densely annotated occupancy labels for training.
In the future, we will explore its self-supervised variant and its application for multi-robot coordination.




{\scriptsize
\bibliographystyle{IEEEtran}
\bibliography{references}
}


\end{document}